\documentclass[letterpaper,10pt,conference]{IEEEtran}
\usepackage[letterpaper]{geometry}
\usepackage{array}
\usepackage{booktabs}
\usepackage{amsmath}
\usepackage{amssymb}
\usepackage{graphicx}

\makeatletter

\providecommand{\tabularnewline}{\\}

\usepackage{url}

\makeatother

\begin{document}

\title{A Compressive Multi-Kernel Method for Privacy-Preserving Machine
Learning}

\author{

\IEEEauthorblockN{Thee Chanyaswad}
\IEEEauthorblockA{Department of \\Electrical Engineeering\\
Princeton University\\
Princeton, New Jersey 08540}
\and
\IEEEauthorblockN{J. Morris Chang}
\IEEEauthorblockA{Department of \\Electrical Engineeering\\
University of South Florida\\
Tampa, Florida 33647}
\and
\IEEEauthorblockN{S.Y. Kung}
\IEEEauthorblockA{Department of \\Electrical Engineeering\\
Princeton University\\
Princeton, New Jersey 08540}

}

\maketitle

\begin{abstract}
As the analytic tools become more powerful, and more data are generated
on a daily basis, the issue of data privacy arises. This leads to
the study of the design of privacy-preserving machine learning algorithms.
Given two objectives, namely, utility maximization and privacy-loss
minimization, this work is based on two previously non-intersecting
regimes \textendash{} \emph{Compressive Privacy} and \emph{multi-kernel
method}. Compressive Privacy is a privacy framework that employs utility-preserving
lossy-encoding scheme to protect the privacy of the data, while multi-kernel
method is a kernel-based machine learning regime that explores the
idea of using multiple kernels for building better predictors. In
relation to the neural-network architecture, multi-kernel method can
be described as a two-hidden-layered network with its width proportional
to the number of kernels.

The compressive multi-kernel method proposed consists of two stages
\textendash{} the compression stage and the multi-kernel stage. The
compression stage follows the Compressive Privacy paradigm to provide
the desired privacy protection. Each kernel matrix is compressed with
a lossy projection matrix derived from the Discriminant Component
Analysis (DCA). The multi-kernel stage uses the \emph{signal-to-noise
ratio (SNR)} score of each kernel to non-uniformly combine multiple
compressive kernels.

The proposed method is evaluated on two mobile-sensing datasets \textendash{}
MHEALTH and HAR \textendash{} where activity recognition is defined
as utility and person identification is defined as privacy. The results
show that the compression regime is successful in privacy preservation
as the privacy classification accuracies are almost at the random-guess
level in all experiments. On the other hand, the novel SNR-based multi-kernel
shows utility classification accuracy improvement upon the state-of-the-art
in both datasets. These results indicate a promising direction for
research in privacy-preserving machine learning.
\end{abstract}

\section{Introduction}

Kernel-based machine learning has been a successful paradigm in various
data analytic applications \cite{RefWorks:33,RefWorks:199,RefWorks:200,RefWorks:268}.
Traditionally, a single kernel is chosen for each system by the user's
expertise and experience. More recent works have investigated the
possibility of using more than one kernel in a system \cite{RefWorks:33,RefWorks:201,RefWorks:202,RefWorks:203,RefWorks:204,RefWorks:205,RefWorks:209}.
The results from these recent works suggest that this multi-kernel
method is attractive in improving the utility of the learning system.
From the perspective of neural network architecture, the multi-kernel
method can be viewed as a two-hidden-layered network with the first
hidden layer corresponds to the kernel feature vector mapping, and
the second hidden layer corresponds to the combination of multiple
kernel feature vectors. However, the learning is only required on
the second hidden layer, as the first first hidden layer is defined
by the kernels. An illustration of the first hidden layer can be found
in \cite{RefWorks:33}. Importantly, the width of the first hidden
layer depends on the number kernels, so higher number of kernels yields
wider networks.

Concurrently, as the data analytical tools become more powerful and
more data are being collected on the daily basis, the concern of data
privacy arises. This is evident from several data leakage incidents
that have been reported \cite{RefWorks:197}. Moreover, the leakage
can even occur when the data are intentionally released with anonymity
\cite{RefWorks:181,RefWorks:182,RefWorks:183,RefWorks:168}. This
shows that security and anonymity may not be sufficient to provide
privacy protection. To this end, a new regime of data privacy preservation
has been proposed to protect privacy at the user end before the data
are shared to the public space. This regime of \emph{Compressive Privacy}
\cite{RefWorks:270,RefWorks:269} aims at modifying the data in a
lossy fashion in order to limit the amount of information on the data
released. 

Naturally, using multiple kernels in a machine learning system increases
the amount of information involved in the learning process. This,
therefore, may provide a challenge in terms of privacy. Although several
works have been carried out on multi-kernel machine learning, they
almost exclusively focus on improving utility of the system. However,
in the current era when privacy concern has become more prevalent,
in order for a multi-kernel machine learning system to be applicable
in the real world, privacy preservation should be incorporated into
the design of the system.

In this work, Compressive Privacy is applied to multi-kernel machine
learning to propose a compressive multi-kernel learning system. The
system consists of two steps \textendash{} the \emph{compression}
step and the \emph{multi-kernel} step. In the compression step, each
kernel is rank-reduced to produce a lossy kernel matrix, while in
the multi-kernel step, multiple lossy kernel matrices are combined
to generate the final multi-kernel matrix to be used in the learning
process. The primary focus of this work is on the classification problem
and two classification problems are defined \textendash{} one for
utility and one for privacy. The design goal is to yield high utility
accuracy, while allowing low privacy accuracy.

The compression step utilizes a powerful dimensionality reduction
framework called Discriminant Component Analysis (DCA) \cite{RefWorks:33,RefWorks:147,RefWorks:216}.
DCA allows the dimensionality reduction to be targeted to a specified
goal, namely, utility classification. More importantly, given $L_{u}$
classes of utility classification, DCA only requires $L_{u}-1$ dimensions
to reach the maximum discriminant power, permitting high compression
when $L_{u}$ is much less than the original dimension of the data.
The multi-kernel step, meanwhile, employs the non-negative linear
combination of kernels, using a novel non-uniform weighting method
based on the \emph{signal-to-noise ratio} (SNR) score of each kernel,
as inspired by the inter-class separability metric presented in \cite{RefWorks:270}.

The proposed method is evaluated on two datasets \textendash{} MHEALTH
\cite{RefWorks:206,RefWorks:207}, and HAR \cite{RefWorks:169}. Both
datasets are related to mobile sensing, of which application is apparent
in the current world of ubiquitous smartphones. For both datasets,
the classification of human activity is defined as utility, whereas
identification of the subject the features are generated from is defined
as privacy. Two experiments are performed on each dataset. The first
experiment uses multiple radial basis function (RBF) kernels with
different gamma values, while the second uses different types of kernels.

The experimental results show that the proposed compressive SNR-based
multi-kernel learning system can improve the utility accuracy when
compared to the compressive single kernel, the compressive uniform
multi-kernel, and the compressive alignment-based multi-kernel learning
systems in all experiments. On the other hand, the privacy accuracy
is reduced to almost random guess in all experiments. The compressive
multi-kernel learning method, therefore, presents a promising framework
for privacy-preserving machine learning applications.

\section{Prior Works}

\subsection{Compressive Privacy}

Compressive Privacy \cite{RefWorks:270,RefWorks:269} is a privacy
paradigm of which the main principle is based on lossy encoding. In
Compressive Privacy, the utility and privacy are known to the system
designer. Consequently, the lossy encoding scheme can be designed
such that it is information-preserving with respect to the utility,
but information-lossy with respect to the privacy. 

Another important principle of Compressive Privacy is that the data
privacy protection should be managed by the data owners. Therefore,
Compressive Privacy defines two different spheres in the world of
data privacy \textendash{} the \emph{private sphere} and the \emph{public
sphere}. The private sphere is where the data owners generate and
manage the lossy encoding of the data, whereas the public sphere is
where the data are available to the authorized users and, at the same
time, to the possible adversary. Therefore, in Compressive Privacy,
data need to be transformed in a privacy-lossy, utility-preserving
way in the private sphere before being transferred to the public sphere.

\subsection{Multi-Kernel Method}

The study of the multi-kernel method arises from the traditional need
to hand-design a kernel for an application, and the idea that combining
multiple kernels should be more powerful than a single kernel. This
has been a problem of interest in the research community both in theoretical,
and application viewpoints. Several works have investigated different
ways of formulating kernel combination including using semidefinite
programming \cite{RefWorks:203}, semi-infinite programming \cite{RefWorks:217},
convex combination \cite{RefWorks:211,RefWorks:212,RefWorks:214},
non-stationary combination \cite{RefWorks:213}, and sparse combination
\cite{RefWorks:215}. At the same time, several applications of the
multi-kernel method have been proposed such as in image analysis \cite{RefWorks:272,RefWorks:274},
neuroimaging \cite{RefWorks:271}, signal processing \cite{RefWorks:275},
and stock market prediction \cite{RefWorks:276}.

Arguably, the most common kernel combination method is the non-negative
linear combination of kernels, which can be viewed as a convex combination
of a finite set of kernels. This multi-kernel is, thus, in the form,
\begin{equation}
\mathbf{K}_{\mu}=\sum_{l=1}^{P}\mu_{l}\mathbf{K}_{l}\label{eq:Kmu}
\end{equation}
where $\mathbf{K}_{l}$ is the base kernel, $P$ is the number of
kernels used, and $\mu_{l}$ is the weight corresponding to each kernel
in the finite set. The design of $\mu_{l}$ has, hence, been the focus
of the research in this area, where traditionally, the uniform weight
has been shown to perform reasonably well \cite{RefWorks:209}, and
more recently, the alignment-based method has been proposed to be
the improvement \cite{RefWorks:201}. For its wealth in past study
in the literature and its simplicity, this work, therefore, follows
this form of the multi-kernel method.

\section{The Proposed Compressive Multi-Kernel Learning Method}

\subsection{Problem Definition}

The tasks considered are classification problems of the utility and
privacy classes. Given $M$-dimensional feature data of $N$ samples,
$\mathbf{X}\in\mathbb{R}^{M\times N}$, and the corresponding utility
label, $\mathbf{y}_{u}$ and privacy label $\mathbf{y}_{p}$, the
goal is to design a system to predict $\mathbf{y}_{u}$ well, but
not $\mathbf{y}_{p}$. The scenario considered is the mobile sensing
application. The utility task is defined to be activity recognition,
while the privacy concern is defined to be the identification of the
data owner.

\subsection{Method Overview}

\begin{figure*}
\begin{minipage}[t]{0.25\paperwidth}%
\begin{flushright}
(a)\includegraphics[scale=0.35]{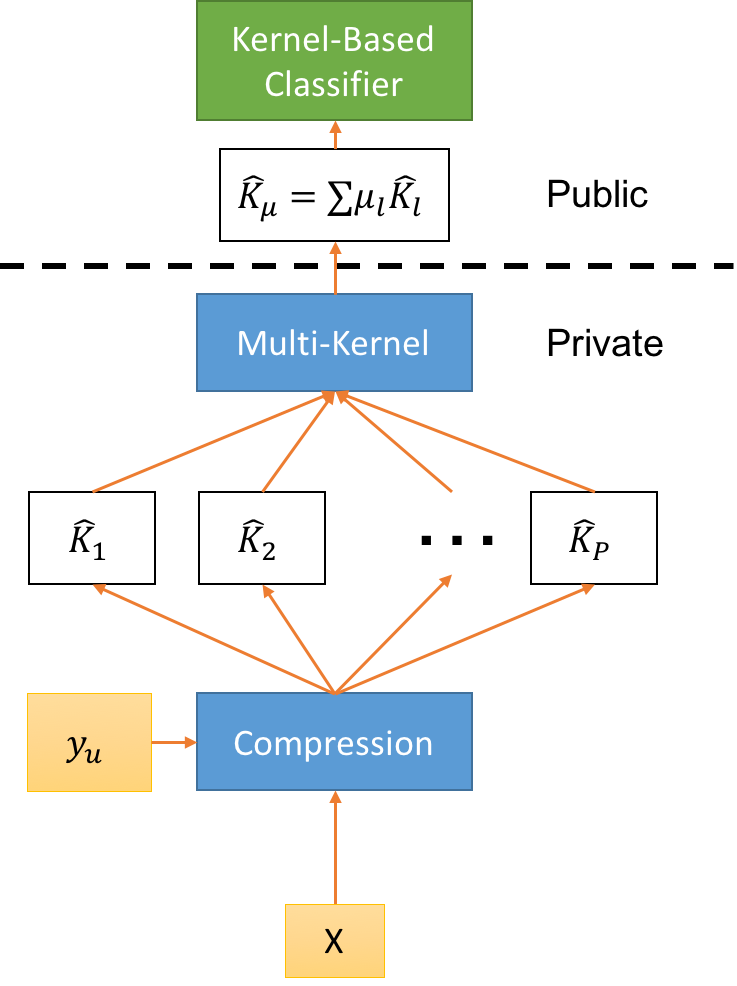}
\par\end{flushright}%
\end{minipage}%
\begin{minipage}[t]{0.56\paperwidth}%
\begin{flushright}
(b)\includegraphics[scale=0.3]{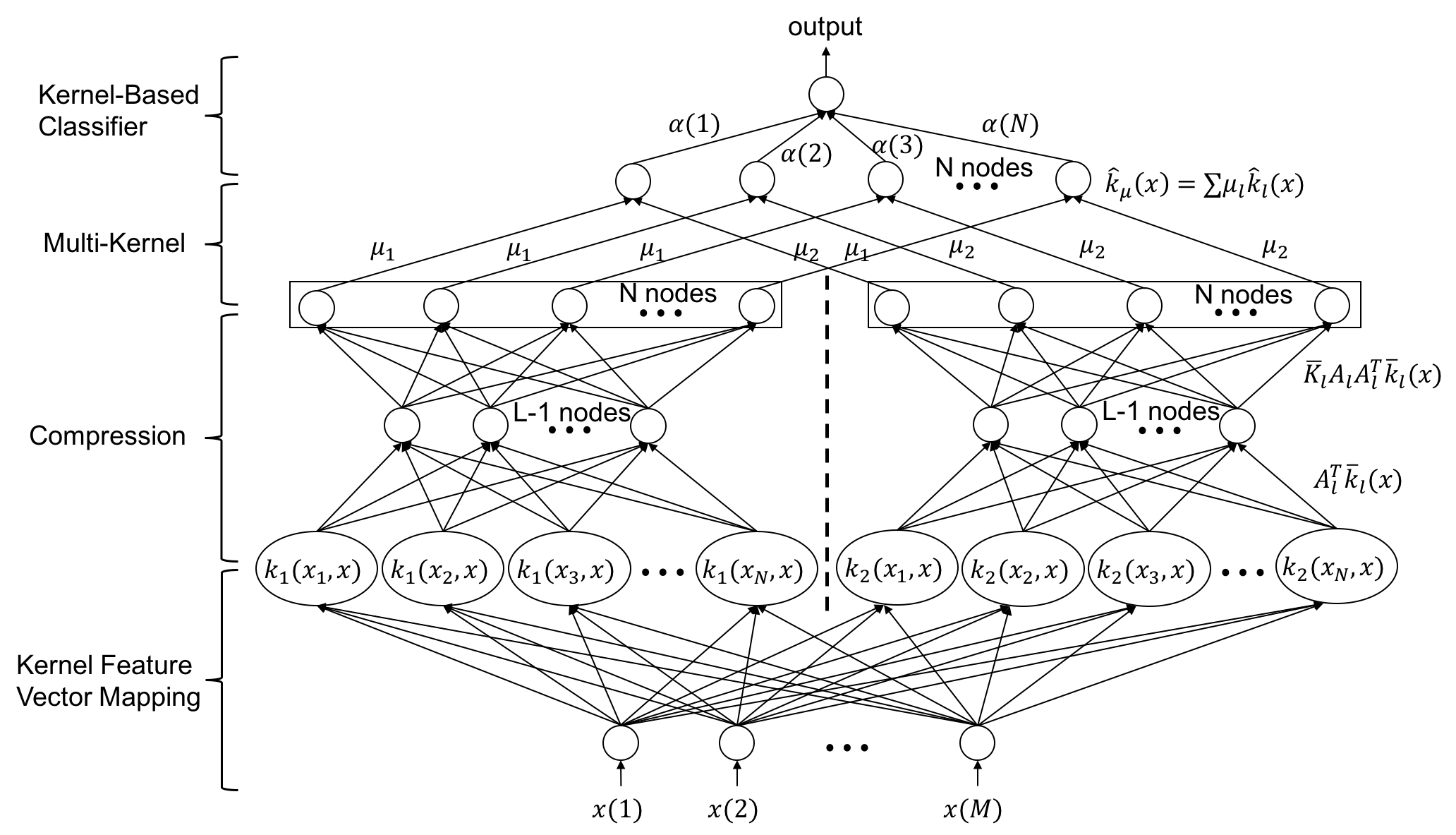}
\par\end{flushright}%
\end{minipage}

\caption{(a) The schematics of the compressive multi-kernel learning system.
The compressive multi-kernel learning phase is considered to be private,
while the classifier phase is consider to be public. (b) A neural
network topology of the compressive multi-kernel learning system with
two kernels. The two kernels are separated in the illustration by
the dashed line. \label{fig:system_schematics}}

\end{figure*}

The proposed compressive multi-kernel learning system consists of
two steps \textendash{} the \emph{compression} step and the \emph{multi-kernel}
step. In the compression step, the training feature data and utility
label are provided and $P$ rank-reduced kernels, $\hat{\mathbf{K}}_{1},\hat{\mathbf{K}}_{2},\ldots,\hat{\mathbf{K}}_{P}$,
are produced. These rank-reduced kernels are then combined in the
multi-kernel step to form the final compressive multi-kernel in the
form $\hat{\mathbf{K}}_{\mu}=\sum_{l=1}^{P}\mu_{l}\hat{\mathbf{K}}_{l}$.
This compressive multi-kernel is provided to the kernel-based classifier.

As common in the framework of Compressive Privacy \cite{RefWorks:270,RefWorks:269},
the private and public spheres are separated. Both compression and
multi-kernel steps are assumed to occur in the private sphere, possibly
by a trusted centralized server, while the classification step is
considered to be in the public sphere. Therefore, the final compressive
multi-kernel $\hat{\mathbf{K}}_{\mu}$ is considered public, whereas
the original feature data, $\mathbf{X}$, are private. Moreover, to
evaluate the system against a strong adversary, the training data,
including the feature data, the utility label and the privacy label,
are assumed to be public, and the adversary is assumed to have the
knowledge of the procedure of the system. The testing data are, on
the other hand, considered private. Figure \ref{fig:system_schematics}(a)
summarizes the system and marks the private and public spheres.

In addition, the proposed compressive multi-kernel learning system
can be visualized as a neural network illustrated in Figure \ref{fig:system_schematics}(b).
The network has four hidden layers. The first hidden layer provides
the kernel feature vector mapping (as described in more detail in
\cite{RefWorks:33}). The second and third hidden layers act as the
compression step, while the the fourth hidden layer serves as the
multi-kernel step. The illustration in Figure \ref{fig:system_schematics}(b)
shows the system with two kernels, as separated by the dashed line.
The last layer in the network is simply the kernel-based classifier
in the system that produces a prediction as the output.

\subsection{The Compression Step}

\subsubsection{Preliminary}

The compression step aims at finding a low-rank approximation of each
kernel matrix $\mathbf{K}_{l}$. The compressive kernel matrix considered
is in the form,
\begin{equation}
\hat{\mathbf{K}}_{l}=\bar{\boldsymbol{\Phi}}^{T}\mathbf{L}\mathbf{L}^{T}\bar{\boldsymbol{\Phi}}\label{eq:kbar_l}
\end{equation}
where $\bar{\boldsymbol{\Phi}}\in\mathbb{R}^{J\times N}$ is the centered
feature data matrix in the intrinsic space as embedded by the kernel
function, and $\mathbf{L}\in\mathbb{R}^{J\times Q};Q<J$ is the \emph{lossy
projection matrix}. In the proposed system, $\mathbf{L}$ is derived
from a powerful dimensionality reduction framework called Discriminant
Component Analysis (DCA) \cite{RefWorks:33,RefWorks:147,RefWorks:216}. 

\subsubsection{Discriminant Component Analysis\label{subsec:Discriminant-Component-Analysis}}

Discriminant Component Analysis (DCA) aims at finding the projection
matrix that maximizes the discriminant power. Given a set of training
feature data and the specified target label $\{\mathbf{X},\mathbf{y}\}$,
DCA utilizes the \emph{between-class scatter matrix} $\mathbf{S}_{B}$,
and the \emph{within-class scatter matrix} $\mathbf{S}_{w}$, defined
by,
\begin{equation}
\mathbf{S}_{B}=\sum_{l=1}^{L}N_{l}[\boldsymbol{\mu}_{l}-\boldsymbol{\mu}][\boldsymbol{\mu}_{l}-\boldsymbol{\mu}]^{T}\label{eq:Sb}
\end{equation}
\begin{equation}
\mathbf{S}_{W}=\sum_{l=1}^{L}\sum_{j=1}^{N_{l}}[\mathbf{x}_{j}^{(l)}-\boldsymbol{\mu}_{l}][\mathbf{x}_{j}^{(l)}-\boldsymbol{\mu}_{l}]^{T}\label{eq:Sw}
\end{equation}
where $L$ is the number of classes, $N_{l}$ is the number of samples
in each class, $\boldsymbol{\mu}_{l}$ is the mean of the samples
in the class, and $\boldsymbol{\mu}$ is the mean of the entire dataset.
It can be shown that the sum of the two scatter matrices equals the
overall scatter matrix of the dataset, namely, 
\begin{equation}
\bar{\mathbf{S}}=\bar{\mathbf{X}}\bar{\mathbf{X}}^{T}=\mathbf{S}_{B}+\mathbf{S}_{W}\label{eq:Sbar}
\end{equation}
 where $\bar{\mathbf{X}}$ is the centered feature data matrix.

From the two scatter matrices, the discriminant power can be defined
as,
\begin{equation}
P(\mathbf{W})=\sum_{i=1}^{Q}\frac{\mathbf{w}_{i}^{T}[\mathbf{S}_{B}+\rho'\mathbf{I}]\mathbf{w}_{i}}{\mathbf{w}_{i}^{T}[\bar{\mathbf{S}}+\rho\mathbf{I}]\mathbf{w}_{i}}\label{eq:Pw_dca}
\end{equation}
where $Q$ is the dimension of the DCA-projected subspace, $\mathbf{w}_{i}$
is each component of the projection matrix, and $\rho$ and $\rho'$
are the ridge parameters. By imposing the canonical orthonormality
that $\mathbf{W}_{DCA}^{T}[\bar{\mathbf{S}}+\rho\mathbf{I}]\mathbf{W}_{DCA}=\mathbf{I}$,
the optimization formulation for DCA can be written as,
\begin{equation}
\mathbf{W}_{DCA}=\underset{\{\mathbf{W}:\mathbf{W}^{T}[\bar{\mathbf{S}}+\rho\mathbf{I}]\mathbf{W}=\mathbf{I}\}}{\arg\max}trace(\mathbf{W}^{T}[\mathbf{S}_{B}+\rho'\mathbf{I}]\mathbf{W})\label{eq:DCA_opt}
\end{equation}
 This optimization can be solved with a generalized eigenvalue problem
solver and the $Q$ eigenvectors with the highest corresponding eigenvalues
are chosen to form the projection matrix.

One important property of DCA is that DCA only requires $L-1$ dimensions
to reach the maximum discriminant power as $\mathbf{S}_{B}$ has rank
of $\leq L-1$. This property, hence, allows very high compression
when $L\ll M$. Therefore, it is appropriate to call the subspace
spanned by the first $L-1$ DCA components as the \emph{signal subspace},
and that spanned by the rest of the components as the \emph{noise
subspace}. These two subspaces are obviously with respect to the specified
target label used to derive the DCA components.

Another important property of DCA is that DCA satisfies the \emph{learning
subspace property} (LSP) \cite{RefWorks:33}, meaning that each component
of DCA can be written as a linear combination of the training feature
data, i.e., $\mathbf{w}_{i}=\bar{\mathbf{\Phi}}\mathbf{\alpha}_{i}$.
Consequently, the DCA formulation can be kernelized, and the Kernel
DCA (KDCA) optimizer is \cite{RefWorks:33,RefWorks:147},
\begin{equation}
\mathbf{A}_{KDCA}=\underset{\{\mathbf{A}:\mathbf{A}^{T}[\bar{\mathbf{K}}^{2}+\rho\bar{\mathbf{K}}]\mathbf{A}=\mathbf{I}\}}{\arg\max}trace(\mathbf{A}^{T}[\mathbf{K}_{B}+\rho'\bar{\mathbf{K}}]\mathbf{A})\label{eq:KDCA_opt}
\end{equation}
where $\mathbf{A}_{KDCA}\in\mathbb{R}^{N\times Q}$ is the projection
matrix in the empirical space, and $\bar{\mathbf{K}}$, $\mathbf{K}_{B}$
and $\mathbf{K}_{W}$ are defined as follows:
\begin{equation}
\bar{\mathbf{K}}=\mathbf{C}_{N}\mathbf{K}\mathbf{C}_{N}\label{eq:Kbar}
\end{equation}
\begin{equation}
\mathbf{K}_{B}=\mathbf{C}_{N}[\sum_{l=1}^{L}N_{l}(\mathbf{k}(\boldsymbol{\mu}_{l})-\mathbf{k}(\boldsymbol{\mu}))(\mathbf{k}(\boldsymbol{\mu}_{l})-\mathbf{k}(\boldsymbol{\mu}))^{T}]\mathbf{C}_{N}\label{eq:Kb}
\end{equation}
\begin{equation}
\mathbf{K}_{W}=\mathbf{C}_{N}[\sum_{l=1}^{L}\sum_{j=1}^{N_{l}}(\mathbf{k}(\mathbf{x}_{j}^{(l)})-\mathbf{k}(\boldsymbol{\mu}_{l}))(\mathbf{k}(\mathbf{x}_{j}^{(l)})-\mathbf{k}(\boldsymbol{\mu}_{l}))^{T}]\mathbf{C}_{N}\label{eq:Kw}
\end{equation}
where
\[
\mathbf{k}(\boldsymbol{\mu}_{l})=\sum_{\mathbf{x}_{j}\in\mathcal{C}_{l}}\frac{\mathbf{k}(\mathbf{x}_{j})}{N_{l}}
\]
\[
\mathbf{k}(\boldsymbol{\mu)}=\sum_{j=1}^{N}\frac{\mathbf{k}(\mathbf{x}_{j})}{N}
\]
and $\mathbf{C}_{N}=[\mathbf{I}-\frac{1}{N}\mathbf{e}\mathbf{e}^{T}];\ \mathbf{e}=[1\ 1\cdots1]^{T}$
is the centering matrix, $\mathbf{K}$ is the kernel matrix, and $\mathbf{k}(\mathbf{x}_{j})$
is the kernel vector. Finally, it can be shown that \cite{RefWorks:33,RefWorks:147}
\begin{equation}
\bar{\mathbf{K}}^{2}=\mathbf{K}_{B}+\mathbf{K}_{W}\label{eq:Kbar2}
\end{equation}

\subsubsection{Compressive Kernel Matrix}

The lossy projection matrix $\mathbf{L}$ is derived from DCA in the
intrinsic space, $\mathbf{W}_{DCA}$. Thus, it also follows the learning
subspace property, i.e., $\mathbf{L}=\bar{\mathbf{\Phi}}\mathbf{A}_{KDCA}$.
This allows the compressive kernel matrix to be derived from the kernel
matrix in the empirical space as follows:
\begin{equation}
\hat{\mathbf{K}}_{l}=\bar{\mathbf{K}}_{l}\mathbf{A}_{l}\mathbf{A}_{l}^{T}\bar{\mathbf{K}}_{l}\label{eq:khat_l_kernel}
\end{equation}
This can be seen simply by substituting $\mathbf{L}$ in Equation
(\ref{eq:kbar_l}) with $\bar{\mathbf{\Phi}}\mathbf{A}_{KDCA}$ and
using the fact that $\bar{\mathbf{K}}_{l}=\bar{\boldsymbol{\Phi}}_{l}^{T}\bar{\boldsymbol{\Phi}}_{l}$.
The centered kernel matrix, $\bar{\mathbf{K}}_{l}$, can also be derived
directly from the non-centered kernel matrix by:
\begin{equation}
\bar{\mathbf{K}}_{l}=\mathbf{C}_{N}\mathbf{K}_{l}\mathbf{C}_{N}\label{eq:Kbar_l}
\end{equation}

Given $\mathbf{A}_{l}\in\mathbb{R}^{N\times Q}$, it can be shown
that rank of $\hat{\mathbf{K}}_{l}$ is $\leq Q$ \cite{RefWorks:208}.
Moreover, as the derivation of $\mathbf{A}_{l}$ requires only the
kernel function corresponding to $\mathbf{K}_{l}$, but not those
corresponding to other $\mathbf{K}_{j}$ used in the multi-kernel
learning system, each $\mathbf{A}_{l}$ can be derived independently
for each kernel function. The choice of $Q$ can also be chosen independently,
as will be apparent in the multi-kernel step.

In the proposed system, the utility label $\mathbf{y}_{u}$ is used
to derive the lossy projection matrix. This is to provide high compression
when the number of utility class, $L_{u}$, is smaller than the original
feature size, and at the same time, retain high classification power
of the kernel matrix. It can be seen from the DCA formulation that,
in fact, all discriminant power of the kernel matrix should remain
when at least $L_{u}-1$ most-discriminant DCA components are used
to form the lossy projection matrix. Thus, the compressive kernel
matrix should provide the win-win situation in terms of privacy-utility
tradeoff, namely, high compression is achieved for privacy, while
maximum discriminant power remains for utility.

\subsection{The Multi-Kernel Step\label{subsec:The-Multi-Kernel-Step}}

\subsubsection{Preliminaries}

The proposed multi-kernel is in the form,
\begin{equation}
\hat{\mathbf{K}}_{\mu}=\sum_{l=1}^{P}\mu_{l}\hat{\mathbf{K}}_{l}\label{eq:khat_mu}
\end{equation}
where $\mu_{l}$ is the weight corresponding to each compressive kernel
matrix. It is common to impose the condition $\mu_{l}\geq0$ to guarantee
that the multi-kernel is a PDS kernel \cite{RefWorks:201,RefWorks:202,RefWorks:203,RefWorks:204,RefWorks:205}.
This multi-kernel can also be written in the intrinsic space by considering
the centered augmented embedded feature data matrix:
\begin{equation}
\bar{\boldsymbol{\Phi}}_{\mu}=\left[\begin{array}{c}
\sqrt{\mu_{1}}\mathbf{L}_{1}^{T}\bar{\boldsymbol{\Phi}}_{1}\\
\sqrt{\mu_{2}}\mathbf{L}_{2}^{T}\bar{\boldsymbol{\Phi}}_{2}\\
\vdots\\
\sqrt{\mu_{P}}\mathbf{L}_{P}^{T}\bar{\boldsymbol{\Phi}}_{P}
\end{array}\right]\label{eq:phibar_mu}
\end{equation}
It can then be shown that,
\[
\hat{\mathbf{K}}_{\mu}=\bar{\boldsymbol{\Phi}}_{\mu}^{T}\bar{\boldsymbol{\Phi}}_{\mu}
\]
\[
\hat{\mathbf{K}}_{\mu}=\sum_{l=1}^{P}\mu_{l}\bar{\boldsymbol{\Phi}}_{l}^{T}\mathbf{L}_{l}\mathbf{L}_{l}^{T}\bar{\boldsymbol{\Phi}}_{l}=\sum_{l=1}^{P}\mu_{l}\hat{\mathbf{K}}_{l}
\]
This shows that the multi-kernel in the form of Equation (\ref{eq:khat_mu})
can be considered as extending the embedded feature in the intrinsic
space according to Equation (\ref{eq:phibar_mu}), and hence, the
additional degrees of freedom. Moreover, there is no overlap between
different embedded features $\bar{\boldsymbol{\Phi}}_{l}$ and its
corresponding lossy projection matrix $\mathbf{L}_{l}$, so each lossy
projection matrix can be designed independently with respect to each
kernel.

However, as the application of privacy is under consideration, extending
the embedded feature space can come at a cost of extra information
shared. Under the regime of Compressive Privacy \cite{RefWorks:270,RefWorks:269},
this is not desirable. Hence, in the design of the multi-kernel step,
the rank of the final multi-kernel should not exceed that of the original
data. Given that each compressive kernel $\hat{\mathbf{K}}_{l}$ has
rank of at most $Q_{l}$, it can be shown that rank of $\hat{\mathbf{K}}_{\mu}$
is $\leq\sum_{l=1}^{P}Q_{l}$ \cite{RefWorks:208}. Thus, to comply
with Compressive Privacy, the compressive multi-kernel has to be designed
such that $\sum_{l=1}^{P}Q_{l}<M$.

\subsubsection{Centering and Normalization}

Following the suggestion in \cite{RefWorks:201}, each compressive
kernel should be centered and normalized before being combined. $\hat{\mathbf{K}}_{l}$
is, in fact, already centered, so only normalization is needed. The
normalization is carried out such that each kernel has trace equals
to one:
\begin{equation}
\hat{\mathbf{K}}[i,j]_{normalized}=\hat{\mathbf{K}}[i,j]/trace(\hat{\mathbf{K}})\label{eq:Khat_normalized}
\end{equation}
where the indices $[i,j]$ indicates the element at row $i^{th}$
and column $j^{th}$.

\subsubsection{Signal-to-Noise Ratio for Kernel Weight Design}

The design of the multi-kernel in Equation (\ref{eq:khat_mu}) involves
the determination of the value of $\mu_{l}$ for each kernel. The
\emph{signal-to-noise ratio (SNR)} in the form of the trace-norm of
the \emph{discriminant matrix}, as motivated by the inter-class separability
metric in Equation (31) of \cite{RefWorks:270}, is proposed as a
new metric to decide the value of $\mu_{l}$. Specifically, the SNR
is defined as,

\begin{equation}
SNR=\left\Vert [\bar{\mathbf{S}}+\rho_{snr}\mathbf{I}]^{-1}\mathbf{S}_{B}\right\Vert _{tr}\label{eq:SNR_intrinsic}
\end{equation}
where $\bar{\mathbf{S}}$ and $\mathbf{S}_{B}$ are defined similarly
to those in DCA (Section \ref{subsec:Discriminant-Component-Analysis}),
$\rho_{snr}$ is the ridge term, and $\left\Vert \cdot\right\Vert _{tr}$
is the trace-norm defined by the sum of the singular values of the
matrix. The discriminant matrix can also be derived directly in the
empirical space by using the counterpart form of each scatter matrix
similar to KDCA as the following:
\begin{equation}
SNR=\left\Vert [\bar{\mathbf{K}}^{2}+\rho_{snr}\mathbf{\bar{K}}]^{-1}\mathbf{K}_{B}\right\Vert _{tr}\label{eq:SNR_K}
\end{equation}
In relation to the compressive kernels, the SNR for each compressive
kernel is,
\begin{equation}
SNR_{l}=\left\Vert [\hat{\mathbf{K}}_{l}^{2}+\rho_{snr}\mathbf{\hat{K}}_{l}]^{-1}\hat{\mathbf{K}}_{B_{l}}\right\Vert _{tr}\label{eq:SNR_l}
\end{equation}
where $\hat{\mathbf{K}}_{B_{l}}=\bar{\mathbf{K}}_{l}\mathbf{A}_{l}\mathbf{A}_{l}^{T}\mathbf{K}_{B_{l}}\mathbf{A}\mathbf{A}^{T}\bar{\mathbf{K}}_{l}$.
Finally, denoting $\boldsymbol{\mu}=[\mu_{1},\mu_{2},\ldots,\mu_{P}]^{T}$
as the kernel weight vector, the value of each $\mu_{l}$ is designed
to be,
\begin{equation}
\mu_{l}=\frac{SNR_{l}}{\left\Vert \boldsymbol{\mu}\right\Vert }\label{eq:mu_l}
\end{equation}
The design of $\boldsymbol{\mu}$ is, therefore, regularized to have
norm of one.

\section{Experimental Results}

\subsection{Datasets}

The proposed system is evaluated on two datasets \textendash{} Mobile
Health (MHEALTH) \cite{RefWorks:206,RefWorks:207}, and Human Activity
Recognition Using Smartphones (HAR) \cite{RefWorks:169}. Both datasets
are related to activity recognition using mobile-sensing features,
and the feature data are collected from multiple individuals, so two
different labels are available \textendash{} activity recognition
label, and person identification label. In all experiments, activity
recognition is defined as the utility classification, whereas person
identification is defined as the privacy classification. Therefore,
the goal of a privacy-preserving classifier is to be able to recognize
the activity label with high accuracy, but identify the person with
low accuracy.

The MHEALTH dataset has 23 features, and 4,800 samples are used for
training, while 900 samples are left out for testing. The utility
label consists of six activity classes, whereas the privacy label
consists of 10 individuals. The HAR dataset has 561 features, and
5,451 samples are kept for training, while 1,080 are used for testing.
The utility label has six activities, and the privacy label has 20
individuals.

\subsection{Experimental Setup}

In all experiments, support vector machine (SVM) is used as the classifier.
The parameters of SVM are selected using cross-validation. To judge
the performance of differential compressive multi-kernel methods fairly,
SVM is used for all methods, and the results from single compressive
kernels are also reported for comparison. Five types of kernels are
used in the experiments and are defined as follows:
\begin{itemize}
\item Linear kernel: $\mathbf{K}_{l}(\mathbf{x}_{i},\mathbf{x}_{j})=\mathbf{x}_{i}^{T}\mathbf{x}_{j}$
\item Polynomial kernel: $\mathbf{K}_{l}(\mathbf{x}_{i},\mathbf{x}_{j})=(\gamma\mathbf{x}_{i}^{T}\mathbf{x}_{j}+c_{0})^{p}$
\item Radial basis function (RBF) kernel: $\mathbf{K}_{l}(\mathbf{x}_{i},\mathbf{x}_{j})=\exp(-\gamma\left\Vert \mathbf{x}_{i}-\mathbf{x}_{j}\right\Vert ^{2})$ 
\item Laplacian kernel: $\mathbf{K}_{l}(\mathbf{x}_{i},\mathbf{x}_{j})=\exp(-\gamma\left\Vert \mathbf{x}_{i}-\mathbf{x}_{j}\right\Vert _{1})$
\item Sigmoid kernel: $\mathbf{K}_{l}(\mathbf{x}_{i},\mathbf{x}_{j})=\tanh(\gamma\mathbf{x}_{i}^{T}\mathbf{x}_{j}+c_{0})$
\end{itemize}
For each dataset, two sets of experiments are performed. The first
experiment is on multiple RBF kernels with various gamma values. The
gamma values are chosen from the set of $a\times10^{b}$, where $a=\{1,3,5\}$
and $b=\{0,1,2,3,4,5\}$, such that, in cross-validation, the single
compressive kernels are sufficiently different in individual performances,
and the multi-kernel performs reasonably well . The second experiment,
on the other hand, uses more than one kind of kernel. Similarly, the
choice of kernels is based on sufficient difference in individual
performances and reasonable multi-kernel performance in the cross-validation
of all above kernels.

For comparison, four methods are investigated in each experiment.
\begin{itemize}
\item The compressive \emph{single kernel}, where only one compressive kernel
is used.
\item The compressive \emph{uniform} multi-kernel, which uses the uniform
weight for the value of $\mu_{l}$. Hence, $\mu_{l}=1/P$ for all
kernels.
\item The compressive \emph{alignment-based multi-kernel}, which uses the
utility-target alignment maximization algorithm according to \cite{RefWorks:201}
on the compressive kernels to determine the value of $\mu_{l}$.
\item The proposed compressive \emph{SNR-based multi-kernel}, which uses
the SNR score described in Section (\ref{subsec:The-Multi-Kernel-Step})
to determine the value of $\mu_{l}$. The value of the ridge term,
$\rho_{snr}$, is chosen to be zero and 0.1 across all experiments.
\end{itemize}
Finally, the details of the choice of kernels in each dataset are
described as follows:

\subsubsection{MHEALTH}

There are six utility classes in the MHEALTH dataset ($L_{u}=6$),
so in order to get the maximum discriminant power, five ($L_{u}-1=5$)
DCA components are used for compression of each kernel in both experiments.
Since $M=23$ for MHEALTH, three kernels are used to keep the compressive
multi-kernel low-rank ($\leq15$). The values of $\rho$ and $\rho'$
of the DCA optimization are kept the same across all kernels in all
experiments with the value of 10 and 0.0001, respectively. The kernels
used in both experiments are as follows:
\begin{itemize}
\item In the first experiment, three RBF kernels with $\gamma=\{0.01,0.03,0.0005\}$.
\item In the second experiment, a polynomial kernel with $\gamma=1.0$,
$p=3$, and $c_{0}=1.0$; an RBF kernel with $\gamma=0.01$; and a
Laplacian kernel with $\gamma=0.1$.
\end{itemize}

\subsubsection{HAR}

There are also six utility classes, so five DCA components are similarly
used for compression of all kernels in both experiments. Since $M=561$
in the HAR dataset, four kernels are used (rank of the compressive
multi-kernel $\leq20$). The values of $\rho$ and $\rho'$ of the
DCA optimization are also kept constant across all kernels at 10 and
0.0001, respectively. The kernels used in both experiments are as
follows:
\begin{itemize}
\item In the first experiment, four RBF kernels with $\gamma=\{10^{-2},10^{-3},10^{-4},10^{-5}\}$.
\item In the second experiment, a linear kernel; an RBF kernel with $\gamma=10^{-3}$;
a Laplacian kernel with $\gamma=0.001$; and a sigmoid kernel with
$\gamma=0.001$ and $c_{0}=1$.
\end{itemize}

\subsection{Results}

\begin{table}
\begin{centering}
\textbf{MHEALTH: Experiment I}
\par\end{centering}
\begin{centering}
\begin{tabular}{>{\centering}p{3.5cm}>{\centering}p{1.8cm}>{\centering}p{1.8cm}}
\toprule 
 & Utility (\%) & Privacy (\%)\tabularnewline
\midrule
\midrule 
Random guess & 16.67 & 10.00\tabularnewline
\midrule 
Compressive single RBF kernel with $\gamma=0.01$ & 78.33 & 15.11\tabularnewline
\midrule 
Compressive single RBF kernel with $\gamma=0.03$ & 76.00 & 10.44\tabularnewline
\midrule 
Compressive single RBF kernel with $\gamma=0.0005$ & 76.33 & 10.22\tabularnewline
\midrule 
Compressive uniform multi-RBF-kernel & 81.11 & 11.00\tabularnewline
\midrule 
Compressive alignment-based multi-RBF-kernel & 84.11 & 10.22\tabularnewline
\midrule 
Compressive SNR-based multi-RBF-kernel $\rho_{snr}=0$ & 82.22 & 13.78\tabularnewline
\midrule 
Compressive SNR-based multi-RBF-kernel $\rho_{snr}=0.1$ & 85.67 & 12.00\tabularnewline
\bottomrule
\end{tabular}
\par\end{centering}
\begin{centering}
\textbf{\\MHEALTH: Experiment II}
\par\end{centering}
\begin{centering}
\begin{tabular}{>{\centering}p{3.5cm}>{\centering}p{1.8cm}>{\centering}p{1.8cm}}
\toprule 
 & Utility (\%) & Privacy (\%)\tabularnewline
\midrule
\midrule 
Random guess & 16.67 & 10.00\tabularnewline
\midrule 
Compressive single polynomial kernel & 76.00 & 15.56\tabularnewline
\midrule 
Compressive single RBF kernel & 78.33 & 15.11\tabularnewline
\midrule 
Compressive single Laplacian kernel & 78.11 & 10.56\tabularnewline
\midrule 
Compressive uniform multi-kernel & 87.11 & 14.22\tabularnewline
\midrule 
Compressive alignment-based multi-kernel & 79.44 & 12.33\tabularnewline
\midrule 
Compressive SNR-based multi-kernel $\rho_{snr}=0$ & 87.33 & 16.00\tabularnewline
\midrule 
Compressive SNR-based multi-kernel $\rho_{snr}=0.1$ & 86.56 & 15.44\tabularnewline
\bottomrule
\end{tabular}
\par\end{centering}
\caption{MHEALTH: utility and privacy classification accuracies \label{tab:mHealth_results}}
\end{table}

\subsubsection{MHEALTH}

Table \ref{tab:mHealth_results} summarizes the results from both
experiments on the MHEALTH dataset. In terms of privacy, both experiments
show that compression is effective in providing privacy as seen by
the fact that the privacy accuracy is almost at the random-guess level
in all compressive single kernels. More importantly, by combining
the compressive single kernels to form the compressive multi-kernel,
the privacy accuracy does not significantly increase and remains almost
at the random-guess level.

In terms of utility, the two compressive SNR-based multi-kernels show
significant improvement in the utility accuracy in both experiments.
Specifically, in the first experiment, the compressive SNR-based multi-kernel
with $\rho_{snr}=0.1$ yields the best result with 7.34\% improvement
over the compressive single kernel; 4.56\% over the compressive uniform
multi-kernel; and 1.56\% over the compressive alignment-based multi-kernel.
Meanwhile, in the second experiment, the compressive SNR-based multi-kernel
with $\rho_{snr}=0$ performs best with 9.00\% higher utility accuracy
than the compressive single kernel; 0.22\% higher than the compressive
uniform multi-kernel; and 7.89\% higher than the compressive alignment-based
multi-kernel.

\begin{table}
\begin{centering}
\textbf{HAR: Experiment I}
\par\end{centering}
\begin{centering}
\begin{tabular}{>{\centering}p{3.5cm}>{\centering}p{1.8cm}>{\centering}p{1.8cm}}
\toprule 
 & Utility (\%) & Privacy (\%)\tabularnewline
\midrule
\midrule 
Random guess & 16.67 & 5.00\tabularnewline
\midrule 
Compressive single RBF kernel with $\gamma=10^{-2}$ & 84.72 & 5.00\tabularnewline
\midrule 
Compressive single RBF kernel with $\gamma=10^{-3}$ & 86.20 & 6.48\tabularnewline
\midrule 
Compressive single RBF kernel with $\gamma=10^{-4}$ & 89.26 & 6.85\tabularnewline
\midrule 
Compressive single RBF kernel with $\gamma=10^{-5}$ & 83.70 & 5.00\tabularnewline
\midrule 
Compressive uniform multi-RBF-kernel & 89.63 & 6.39\tabularnewline
\midrule 
Compressive alignment-based multi-RBF-kernel & 90.19 & 5.00\tabularnewline
\midrule 
Compressive SNR-based multi-RBF-kernel $\rho_{snr}=0$ & 89.35 & 6.85\tabularnewline
\midrule 
Compressive SNR-based multi-RBF-kernel $\rho_{snr}=0.1$ & 90.93 & 5.00\tabularnewline
\bottomrule
\end{tabular}
\par\end{centering}
\begin{centering}
\textbf{\\HAR: Experiment II}
\par\end{centering}
\begin{centering}
\begin{tabular}{>{\centering}p{3.5cm}>{\centering}p{1.8cm}>{\centering}p{1.8cm}}
\toprule 
 & Utility (\%) & Privacy (\%)\tabularnewline
\midrule
\midrule 
Random guess & 16.67 & 5.00\tabularnewline
\midrule 
Compressive single linear kernel & 51.02 & 5.19\tabularnewline
\midrule 
Compressive single RBF kernel & 86.20 & 6.48\tabularnewline
\midrule 
Compressive single Laplacian kernel & 90.83 & 5.00\tabularnewline
\midrule 
Compressive single sigmoid kernel & 82.59 & 7.04\tabularnewline
\midrule 
Compressive uniform multi-kernel & 90.65 & 6.57\tabularnewline
\midrule 
Compressive alignment-based multi-kernel & 91.30 & 6.57\tabularnewline
\midrule 
Compressive SNR-based multi-kernel $\rho_{snr}=0$ & 89.35 & 6.85\tabularnewline
\midrule 
Compressive SNR-based multi-kernel $\rho_{snr}=0.1$ & 91.39 & 5.00\tabularnewline
\bottomrule
\end{tabular}
\par\end{centering}
\caption{ HAR: utility and privacy classification accuracies\label{tab:HAR_results}}
\end{table}

\subsubsection{HAR}

Table \ref{tab:HAR_results} summarizes the results from both experiments
on the HAR dataset. In terms of privacy, both experiments show that
compression is very effective in providing privacy as the privacy
accuracy is equivalent or close to random guess in all compressive
single kernels. Similarly, by combining the compressive single kernels
to form the compressive multi-kernel, the privacy accuracy does not
increase and remains at or close to random guess. 

In terms of utility, the compressive multi-kernel, again, shows improvement
in both experiments on the HAR dataset. The compressive SNR-based
multi-kernel with $\rho_{snr}=0.1$, specifically, yields the best
performance in both experiments by improving upon the compressive
single kernel by at least 1.67\% and 0.56\%; upon the compressive
uniform multi-kernel by 1.30\% and 0.74\%; and upon the compressive
alignment-based multi-kernel in both experiments by 0.74\% and 0.09\%. 

\section{Discussion and Future Works}

\subsection{Compression for Privacy}

The results from the evaluation confirm that the compression step
indeed provides privacy protection as the privacy classification accuracy
is around random guess in all experiments. This is true even when
multiple compressive kernels are used. This success in privacy-preservation
is achievable from the fact that large amount of information is loss
during the compression step.

The compression step is actually general as other dimensionality reduction
techniques can also be used. Future work can be carried out on studying
the effect on the choice of the compression method. Some possible
ideas for future work are the following:
\begin{itemize}
\item Principal Component Analysis (PCA) may be used to maximize data reconstruction
quality when the utility and privacy labels are not known in advance.
\item Desensitized RDCA method \cite{RefWorks:269} may be used when the
privacy label is known at the time of compression but the utility
label is not.
\item Utility-to-privacy ratio \cite{RefWorks:210,RefWorks:270} maybe used
when both utility and privacy labels are known.
\end{itemize}

\subsection{Multi-Kernel for Utility (and Privacy)}

In the current system, the multi-kernel step is primarily for utility
improvement. The results from the evaluation conform with the goal.
Especially, the SNR-based kernel weight score proposed is shown to
be fairly effective and improving upon the state-of-the-art. Nevertheless,
the current work primarily focuses on using the multi-kernel weight
optimization for enhancing utility only, but as the application of
privacy-preserving machine learning involves maximizing utility, while
minimizing privacy, it is interesting to seek an alternative optimization
criterion that fits this objective. 

Fortunately, the SNR score can be extended to such setting by considering
the utility-to-privacy ratio similar to that in \cite{RefWorks:210,RefWorks:270}.
By utilizing both the utility label $\mathbf{y}_{u}$ and privacy
label $\mathbf{y}_{p}$, the \emph{utility-to-privacy ratio} can be
defined as,
\begin{equation}
UPR=\left\Vert [\mathbf{S}_{B_{P}}+\rho_{upr}\mathbf{I}]^{-1}\mathbf{S}_{B_{U}}\right\Vert _{tr}\label{eq:UPR_intrinsic}
\end{equation}
where $\mathbf{S}_{B_{U}}$ is the between-class scatter matrix derived
from $\mathbf{y}_{u}$, $\mathbf{S}_{B_{P}}$ is the between-class
scatter matrix derived from $\mathbf{y}_{p}$, and $\rho_{upr}$ is,
again, the ridge term. Naturally, the utility-to-privacy ratio can
also be derived in the empirical space:
\begin{equation}
UPR=\left\Vert [\mathbf{K}_{B_{P}}+\rho_{upr}\mathbf{\bar{K}}]^{-1}\mathbf{K}_{B_{U}}\right\Vert _{tr}\label{eq:UPR_K}
\end{equation}
where $\mathbf{K}_{B_{U}}$ is the counterpart of $\mathbf{S}_{B_{U}}$
and $\mathbf{K}_{B_{P}}$ is the counterpart of $\mathbf{S}_{B_{P}}$
in the empirical space.

Alternatively, the alignment approach in \cite{RefWorks:202,RefWorks:203}
can also be extended to consider both utility and privacy together
by defining the alignments between the multi-kernel matrix and the
targeted utility and privacy labels, respectively, as:
\begin{equation}
A_{u}(\hat{\mathbf{K}}_{\mu},\mathbf{y}_{u}\mathbf{y}_{u}^{T})=\bigl\langle\hat{\mathbf{K}}_{\mu},\mathbf{y}_{u}\mathbf{y}_{u}^{T}\bigr\rangle_{F}/\bigl\Vert\hat{\mathbf{K}}_{\mu}\bigr\Vert_{F}\label{eq:alignment_u}
\end{equation}
\begin{equation}
A_{p}(\hat{\mathbf{K}}_{\mu},\mathbf{y}_{p}\mathbf{y}_{p}^{T})=\bigl\langle\hat{\mathbf{K}}_{\mu},\mathbf{y}_{p}\mathbf{y}_{p}^{T}\bigr\rangle_{F}/\bigl\Vert\hat{\mathbf{K}}_{\mu}\bigr\Vert_{F}\label{eq:alignment_p}
\end{equation}
Then, following the line of thought in the \emph{Differential Utility/Cost
Analysis} (DUCA) formulation \cite{RefWorks:270}, the utility-to-privacy
ratio can be maximized as the following:
\[
\boldsymbol{\mu}=\underset{\boldsymbol{\mu}:\left\Vert \boldsymbol{\mu}\right\Vert =1,\boldsymbol{\mu}\geq0}{\arg\max}\frac{A_{u}(\hat{\mathbf{K}}_{\mu},\mathbf{y}_{u}\mathbf{y}_{u}^{T})}{A_{p}(\hat{\mathbf{K}}_{\mu},\mathbf{y}_{p}\mathbf{y}_{p}^{T})}=\underset{\boldsymbol{\mu}:\left\Vert \boldsymbol{\mu}\right\Vert =1,\boldsymbol{\mu}\geq0}{\arg\max}\frac{\bigl\langle\hat{\mathbf{K}}_{\mu},\mathbf{y}_{u}\mathbf{y}_{u}^{T}\bigr\rangle_{F}}{\bigl\langle\hat{\mathbf{K}}_{\mu},\mathbf{y}_{p}\mathbf{y}_{p}^{T}\bigr\rangle_{F}}
\]
This optimization criterion can be formulated into a simple quadratic
programming problem. First, defining the following two vectors:
\[
\mathbf{a}_{u}=[\mathbf{y}_{u}^{T}\hat{\mathbf{K}}_{1}\mathbf{y}_{u},\ \mathbf{y}_{u}^{T}\hat{\mathbf{K}}_{2}\mathbf{y}_{u},\ldots,\ \mathbf{y}_{u}^{T}\hat{\mathbf{K}}_{P}\mathbf{y}_{u}]^{T}
\]
\[
\mathbf{a}_{p}=[\mathbf{y}_{p}^{T}\hat{\mathbf{K}}_{1}\mathbf{y}_{p},\ \mathbf{y}_{p}^{T}\hat{\mathbf{K}}_{2}\mathbf{y}_{p},\ldots,\ \mathbf{y}_{p}^{T}\hat{\mathbf{K}}_{P}\mathbf{y}_{p}]^{T}
\]
Then, it can be shown that the solution $\boldsymbol{\mu}^{*}$ is
given by $\boldsymbol{\mu}^{*}=\mathbf{v}^{*}/\left\Vert \mathbf{v}^{*}\right\Vert $,
where $\mathbf{v}^{*}$ is the solution of the following quadratic program:
\begin{equation}
\min_{\mathbf{v}\geq0}\mathbf{v}^{T}\mathbf{a}_{p}\mathbf{a}_{p}^{T}\mathbf{v}-2\mathbf{v}^{T}\mathbf{a}_{u}\label{eq:qp_opt}
\end{equation}
The proof of this proposition directly follows the proof of the alignment
maximization algorithm in Proposition 3 of \cite{RefWorks:202}. 

These two optimization criteria will be an interesting formulation
for future study, especially for the case when compression alone is
not sufficient for privacy protection.

\subsection{Multiple Utility Labels}

One advantage of the multi-kernel in the form of Equation (\ref{eq:khat_mu})
is the independence among the kernels, which, as a result, allows
each kernel to be designed for different purposes. This lends itself
nicely to the application when there are more than one utility label.
For example, different kinds of kernel can be chosen for each utility
label, and the multi-kernel should then be useful for more than one
utility simultaneously. Alternatively, under the compressive kernel
scheme, the DCA-derived lossy projection matrix used for kernel compression
can be derived from DCA with respect to different utility labels for
different kernels. This would allow the multi-kernel to be applicable
to more than one utility label and possible reduce conflict in the
compression design.

\section{Conclusion}

In addressing the challenge problem of privacy-preserving machine
learning, this work is built upon two regimes \textendash{} Compressive
Privacy for privacy protection, and multi-kernel method for utility
maximization. The tasks considered are the classifications of the
utility and privacy labels. The proposed compressive multi-kernel
method consists of two steps \textendash{} compression and multi-kernel.
In the compression step, each kernel is compressed using the DCA lossy
projection matrix. In the multi-kernel step, multiple compressive
kernels are combined based on the proposed SNR score.

The proposed method is evaluated on the utility and privacy classification
tasks using two datasets \textendash{} MHEALTH and HAR. The results
show that the compression step effectively provides privacy protection
by reducing the privacy accuracy to almost random guess in all experiments.
On the other hand, the compressive multi-kernel shows improvement
in the utility accuracy over the single compressive kernels in both
datasets. More importantly, the proposed SNR-based multi-kernel shows
enhancing performance in comparison to the previous uniform and alignment-based
methods. Overall, this work offers a promising direction to the challenging
problem of privacy-preserving machine learning.

\section*{Acknowledgement}

This material is based upon work supported in part by the Brandeis
Program of the Defense Advanced Research Project Agency (DARPA) and
Space and Naval Warfare System Center Pacific (SSC Pacific) under
Contract No. 66001-15-C-4068.



\bibliographystyle{IEEEtran}
\bibliography{reference_short}

\begin{thebibliography}{10}
\providecommand{\url}[1]{#1}
\csname url@samestyle\endcsname
\providecommand{\newblock}{\relax}
\providecommand{\bibinfo}[2]{#2}
\providecommand{\BIBentrySTDinterwordspacing}{\spaceskip=0pt\relax}
\providecommand{\BIBentryALTinterwordstretchfactor}{4}
\providecommand{\BIBentryALTinterwordspacing}{\spaceskip=\fontdimen2\font plus
\BIBentryALTinterwordstretchfactor\fontdimen3\font minus
  \fontdimen4\font\relax}
\providecommand{\BIBforeignlanguage}[2]{{%
\expandafter\ifx\csname l@#1\endcsname\relax
\typeout{** WARNING: IEEEtran.bst: No hyphenation pattern has been}%
\typeout{** loaded for the language `#1'. Using the pattern for}%
\typeout{** the default language instead.}%
\else
\language=\csname l@#1\endcsname
\fi
#2}}
\providecommand{\BIBdecl}{\relax}
\BIBdecl

\bibitem{RefWorks:33}
S.~Y. Kung, \emph{Kernel Methods and Machine Learning}.\hskip 1em plus 0.5em
  minus 0.4em\relax Cambridge, UK: Cambridge University Press, 2014.

\bibitem{RefWorks:199}
B.~Scholkopf and A.~J. Smola, \emph{Learning with kernels: support vector
  machines, regularization, optimization, and beyond}.\hskip 1em plus 0.5em
  minus 0.4em\relax MIT press, 2002.

\bibitem{RefWorks:200}
J.~Shawe-Taylor and N.~Cristianini, \emph{Kernel methods for pattern
  analysis}.\hskip 1em plus 0.5em minus 0.4em\relax Cambridge university press,
  2004.

\bibitem{RefWorks:268}
K.~Xu, H.~Yue, L.~Guo, Y.~Guo, and Y.~Fang, ``Privacy-preserving machine
  learning algorithms for big data systems,'' in \emph{ICDCS 2015}.\hskip 1em
  plus 0.5em minus 0.4em\relax IEEE, 2015, pp. 318--327.

\bibitem{RefWorks:201}
C.~Cortes, M.~Mohri, and A.~Rostamizadeh, ``Two-stage learning kernel
  algorithms,'' in \emph{Proceedings of the 27th ICML}, 2010.

\bibitem{RefWorks:202}
------, ``Learning non-linear combinations of kernels,'' in \emph{Advances in
  neural information processing systems}, 2009, pp. 396--404.

\bibitem{RefWorks:203}
G.~R. Lanckriet, N.~Cristianini, P.~Bartlett, L.~E. Ghaoui, and M.~I. Jordan,
  ``Learning the kernel matrix with semidefinite programming,'' \emph{Journal
  of Machine learning research}, vol.~5, no. Jan, pp. 27--72, 2004.

\bibitem{RefWorks:204}
C.~Cortes, M.~Mohri, and A.~Rostamizadeh, ``L2 regularization for learning
  kernels,'' in \emph{Proceedings of the 25th UAI}.\hskip 1em plus 0.5em minus
  0.4em\relax AUAI Press, 2009.

\bibitem{RefWorks:205}
F.~R. Bach, G.~R. Lanckriet, and M.~I. Jordan, ``Multiple kernel learning,
  conic duality, and the smo algorithm,'' in \emph{Proceedings of the 21st
  ICML}.\hskip 1em plus 0.5em minus 0.4em\relax ACM, 2004, p.~6.

\bibitem{RefWorks:209}
C.~Cortes, ``Can learning kernels help performance,'' in \emph{Invited talk at
  ICML 2009}, 2009.

\bibitem{RefWorks:197}
P.~R. Clearinghouse, ``Chronology of data breaches,''
  \url{http://www.privacyrights.org/data-breach}, 2016.

\bibitem{RefWorks:181}
A.~Narayanan, H.~Paskov, N.~Z. Gong, J.~Bethencourt, E.~Stefanov, E.~C.~R.
  Shin, and D.~Song, ``On the feasibility of internet-scale author
  identification,'' in \emph{2012 IEEE S\&P}.\hskip 1em plus 0.5em minus
  0.4em\relax IEEE, 2012, pp. 300--314.

\bibitem{RefWorks:182}
J.~A. Calandrino, A.~Kilzer, A.~Narayanan, E.~W. Felten, and V.~Shmatikov,
  ``"you might also like:" privacy risks of collaborative filtering,'' in
  \emph{2011 IEEE S\&P}.\hskip 1em plus 0.5em minus 0.4em\relax IEEE, 2011, pp.
  231--246.

\bibitem{RefWorks:183}
M.~Barbaro and T.~Z. Jr., ``A face is exposed for aol searcher no. 4417749,''
  \url{http://www.nytimes.com/2006/08/09/technology/09aol.html}, Aug 9, 2006
  2006.

\bibitem{RefWorks:168}
A.~Narayanan and V.~Shmatikov, ``Robust de-anonymization of large sparse
  datasets,'' in \emph{IEEE S\&P 2008}.\hskip 1em plus 0.5em minus 0.4em\relax
  IEEE, 2008, pp. 111--125.

\bibitem{RefWorks:270}
S.~Kung, ``Compressive privacy: From information\/estimation theory to machine
  learning [lecture notes],'' \emph{IEEE Signal Processing Magazine}, vol.~34,
  no.~1, pp. 94--112, 2017.

\bibitem{RefWorks:269}
S.~Kung, T.~Chanyaswad, J.~M. Chang, and P.~Wu, ``Collaborative {PCA/DCA}
  learning methods for compressive privacy,'' \emph{ACM
  Trans.Embed.Comput.Syst}.

\bibitem{RefWorks:147}
S.-Y. Kung, ``Discriminant component analysis for privacy protection and
  visualization of big data,'' \emph{Multimedia Tools and Applications}, pp.
  1--36, 2015.

\bibitem{RefWorks:216}
T.~Chanyaswad, J.~M. Chang, P.~Mittal, and S.~Kung, ``Discriminant-component
  eigenfaces for privacy-preserving face recognition,'' in \emph{MLSP
  2016}.\hskip 1em plus 0.5em minus 0.4em\relax IEEE, 2016, pp. 1--6.

\bibitem{RefWorks:206}
O.~Banos, R.~Garcia, J.~A. Holgado-Terriza, M.~Damas, H.~Pomares, I.~Rojas,
  A.~Saez, and C.~Villalonga, ``mhealthdroid: a novel framework for agile
  development of mobile health applications,'' in \emph{International Workshop
  on Ambient Assisted Living}.\hskip 1em plus 0.5em minus 0.4em\relax Springer,
  2014, pp. 91--98.

\bibitem{RefWorks:207}
O.~Banos, C.~Villalonga, R.~Garcia, A.~Saez, M.~Damas, J.~A. Holgado-Terriza,
  S.~Lee, H.~Pomares, and I.~Rojas, ``Design, implementation and validation of
  a novel open framework for agile development of mobile health applications,''
  \emph{Biomedical engineering online}, vol.~14, no.~2, p.~1, 2015.

\bibitem{RefWorks:169}
D.~Anguita, A.~Ghio, L.~Oneto, X.~Parra, and J.~L. Reyes-Ortiz, ``A public
  domain dataset for human activity recognition using smartphones.'' in
  \emph{ESANN}, 2013.

\bibitem{RefWorks:217}
P.~Gehler and S.~Nowozin, ``Infinite kernel learning,'' in \emph{NIPS Workshop
  on Kernel Learning: Automatic Selection of Optimal Kernels}, 2008.

\bibitem{RefWorks:211}
C.~A. Micchelli and M.~Pontil, ``Learning the kernel function via
  regularization,'' \emph{Journal of Machine Learning Research}, vol.~6, no.
  Jul, pp. 1099--1125, 2005.

\bibitem{RefWorks:212}
A.~Argyriou, C.~A. Micchelli, and M.~Pontil, ``Learning convex combinations of
  continuously parameterized basic kernels,'' in \emph{International Conference
  on Computational Learning Theory}.\hskip 1em plus 0.5em minus 0.4em\relax
  Springer, 2005, pp. 338--352.

\bibitem{RefWorks:214}
A.~Zien and C.~S. Ong, ``Multiclass multiple kernel learning,'' in
  \emph{Proceedings of the 24th ICML}.\hskip 1em plus 0.5em minus 0.4em\relax
  ACM, 2007, pp. 1191--1198.

\bibitem{RefWorks:213}
D.~P. Lewis, T.~Jebara, and W.~S. Noble, ``Nonstationary kernel combination,''
  in \emph{Proceedings of the 23rd ICML}.\hskip 1em plus 0.5em minus
  0.4em\relax ACM, 2006.

\bibitem{RefWorks:215}
F.~R. Bach, ``Exploring large feature spaces with hierarchical multiple kernel
  learning,'' in \emph{Advances in neural information processing systems},
  2009, pp. 105--112.

\bibitem{RefWorks:272}
Y.-Y. Lin, T.-L. Liu, and C.-S. Fuh, ``Multiple kernel learning for
  dimensionality reduction,'' \emph{IEEE Transactions on Pattern Analysis and
  Machine Intelligence}, vol.~33, no.~6, pp. 1147--1160, 2011.

\bibitem{RefWorks:274}
S.~S. Bucak, R.~Jin, and A.~K. Jain, ``Multiple kernel learning for visual
  object recognition: A review,'' \emph{IEEE Transactions on Pattern Analysis
  and Machine Intelligence}, vol.~36, no.~7, pp. 1354--1369, 2014.

\bibitem{RefWorks:271}
C.~Hinrichs, V.~Singh, J.~Peng, and S.~Johnson, ``Q-mkl: Matrix-induced
  regularization in multi-kernel learning with applications to neuroimaging,''
  in \emph{Advances in neural information processing systems}, 2012, pp.
  1421--1429.

\bibitem{RefWorks:275}
N.~Subrahmanya and Y.~C. Shin, ``Sparse multiple kernel learning for signal
  processing applications,'' \emph{IEEE Transactions on Pattern Analysis and
  Machine Intelligence}, vol.~32, no.~5, pp. 788--798, 2010.

\bibitem{RefWorks:276}
F.~Wang, L.~Liu, and C.~Dou, ``Stock market volatility prediction: a
  service-oriented multi-kernel learning approach,'' in \emph{SCC 2012}.\hskip
  1em plus 0.5em minus 0.4em\relax IEEE, 2012, pp. 49--56.

\bibitem{RefWorks:208}
R.~A. Horn and C.~R. Johnson, \emph{Matrix analysis}.\hskip 1em plus 0.5em
  minus 0.4em\relax Cambridge university press, 2012.

\bibitem{RefWorks:210}
K.~Diamantaras and S.-Y. Kung, ``Data privacy protection by kernel subspace
  projection and generalized eigenvalue decomposition,'' in \emph{MLSP
  2016}.\hskip 1em plus 0.5em minus 0.4em\relax IEEE, 2016, pp. 1--6.

\end{thebibliography}

\end{document}